\documentclass[letterpaper, 10 pt, conference]{ieeeconf}  

\IEEEoverridecommandlockouts                              

\let\labelindent\relax         

\usepackage[normalem]{ulem}                                        
\usepackage{marginnote}

\usepackage[textwidth=10ex,colorinlistoftodos]{todonotes}

\colorlet{mh}{red}
\colorlet{fwu}{red}
\colorlet{kchen}{blue}
\colorlet{zshen}{green}
\colorlet{shaddadin}{purple}
\colorlet{iperez}{cyan}
\colorlet{schneider}{magenta}
\colorlet{lchen}{teal}

\usepackage[inline]{enumitem}
\setlist{nolistsep}

\usepackage{graphics} 
\usepackage{epsfig} 
\usepackage{mathptmx} 
\usepackage{times} 
\usepackage{amsmath} 

\usepackage{amssymb}  
\usepackage{bm}
\usepackage{algorithm}
\usepackage[noend]{algpseudocode}
\usepackage{physics}
\usepackage{xcolor}
\usepackage{tikz}
\usetikzlibrary{fit,
                positioning}
\usetikzlibrary{calc} 
\usetikzlibrary{positioning}
\usepackage{balance}
\usepackage{soul}
\usetikzlibrary{arrows}
\usepackage{cite}
\usepackage{url}
\usepackage{multirow}
\usepackage{booktabs} 
\usepackage{setspace}
\usepackage{listings}
\usepackage{hyperref}

\usepackage{setspace}
\definecolor{commentgreen}{rgb}{0.0, 0.5, 0.0} 
\newcommand{\smalltt}[1]{{\small \texttt{#1}}}
\newcommand{\ftnotett}[1]{{\scriptsize \texttt{#1}}}

\title{\LARGE

LEMMo-Plan: \textbf{L}LM-\textbf{E}nhanced Learning from \textbf{M}ulti-\textbf{Mo}dal Demonstration for \textbf{Plan}ning Sequential Contact-Rich Manipulation Tasks
}

\author{Kejia Chen$^{1^*}$, Zheng Shen$^{{1, 2}^*}$, Yue Zhang$^{1}$, Lingyun Chen$^{1}$,\\ Fan Wu$^{1\dagger}$, Zhenshan Bing$^{1,3}$, Sami Haddadin$^{1,4}$,  Alois Knoll$^{1}$ 
\thanks{\textsuperscript{*} Equal contribution.}
\thanks{$^\dagger$ Corresponding author: Fan Wu ({\tt\small f.wu@tum.de}).}
\thanks{$^{1}$ School of Computation, Information and Technology, Technical University of Munich, Germany.}
\thanks{$^{2}$ Aalto University, Espoo, Finland.}
\thanks{$^{3}$ State Key Laboratory for Novel Software Technology and the School of Science and Technology, Nanjing University (Suzhou Campus), China.}
\thanks{$^{4}$ Mohamed Bin Zayed University of Artificial Intelligence, Abu Dhabi, UAE.}
}

\begin{document}

\maketitle

\begin{abstract}
Large Language Models (LLMs) have gained popularity in task planning for long-horizon manipulation tasks. 
To enhance the validity of LLM-generated plans, visual demonstrations and online videos have been widely employed to guide the planning process. 
However, for manipulation tasks involving subtle movements but rich contact interactions, visual perception alone may be insufficient for the LLM to fully interpret the demonstration. 
Additionally, visual data provides limited information on force-related parameters and conditions, which are crucial for effective execution on real robots.

In this paper, we introduce LEMMo-Plan, an in-context learning framework that incorporates tactile and force-torque information from human demonstrations to enhance LLMs' ability to generate plans for new task scenarios. 
We propose a bootstrapped reasoning pipeline that sequentially integrates each modality into a comprehensive task plan. 
This task plan is then used as a reference for planning in new task configurations.
Real-world experiments on two different sequential manipulation tasks demonstrate the effectiveness of our framework in improving LLMs' understanding of multi-modal demonstrations and enhancing the overall planning performance. More materials are available on our project website: ~\href{https://lemmo-plan.github.io/LEMMo-Plan/}{lemmo-plan.github.io/LEMMo-Plan/}.



\end{abstract}

\section{INTRODUCTION}






Recent advances in Large Language Models (LLMs)~\cite{openai2024gpt4technicalreport,brown2020GPT3} and Visual-Language Models (VLMs)~\cite{radford2021CLIP} have triggered a paradigm shift in the domain of long-horizon task planning~\cite{meng2025preserving}. 
By leveraging the semantic understanding and reasoning capabilities of LLMs, immense progress has been made towards developing task-agnostic high-level task planner~\cite{huang2024understanding,10.5555/3618408.3618748}.
The majority of prior works in LLM-based task planning use in-context learning techniques where carefully designed examples are employed to generate high-level task plans~\cite{song2023llmplanner,liang2023code,xie2023translating,zhou2024isr}, and leverage visual information for skill grounding. 


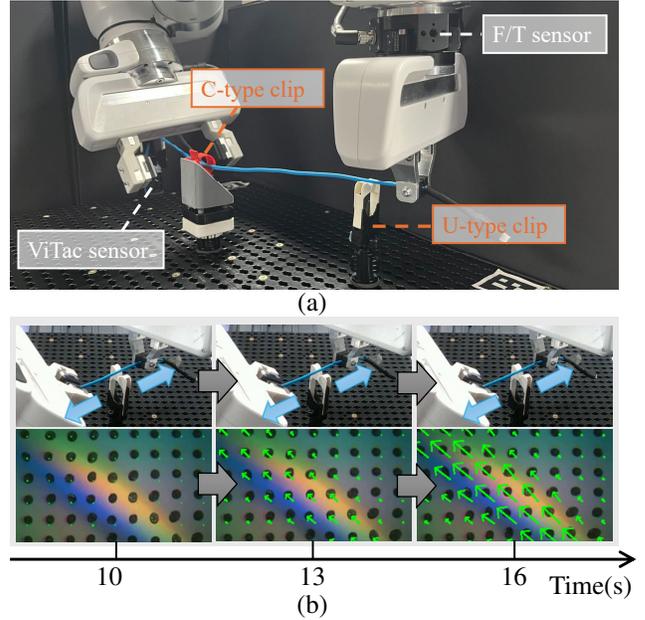
\begin{figure}[t!]
    \centering
    \begin{tikzpicture}
    \node [inner sep=0pt] (russell) at (0,0)
    {\includegraphics[width=0.48\textwidth]{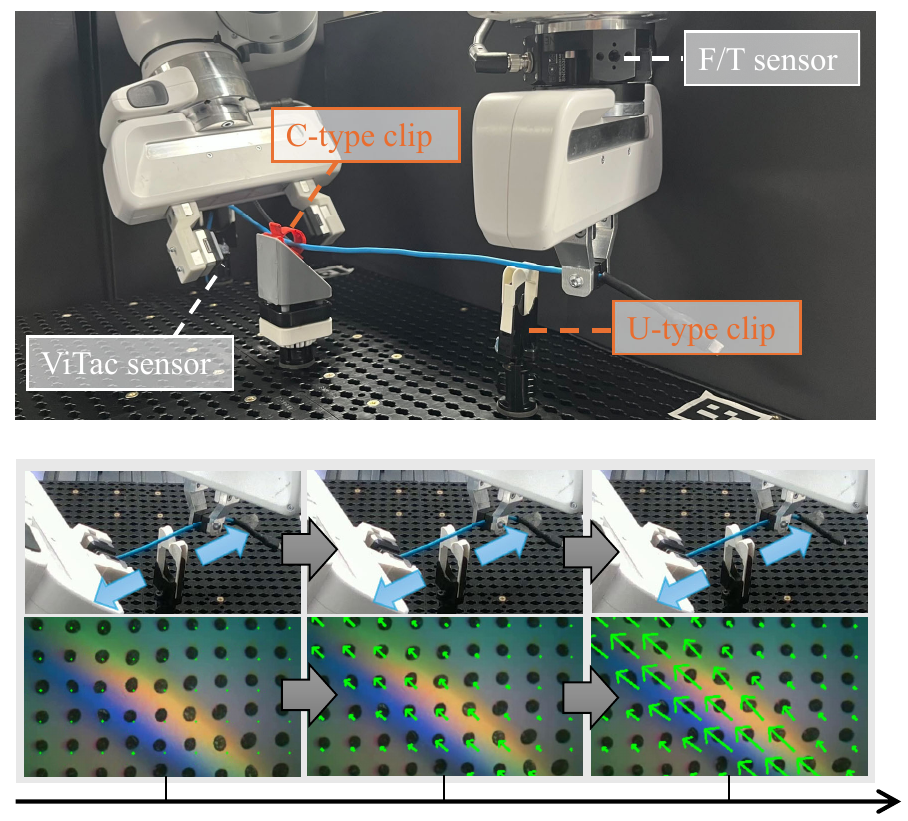}};
    \node at (3.6,-4.05) {Time(s)};
    \node at (-2.8,-3.9) {10};
    \node at (-.1,-3.9) {13};
    \node at (2.58,-3.9) {16};

    \node[text=white] at (-3.85,3.54) {(a)};
    \node[text=white] at (-3.8,-3.2) {(b)};

    
    \end{tikzpicture}
     \vspace{-2.0em} 
    \caption{Cable Manipulation with Two Robots. (a) Setup for cable mounting demonstration. The human operator controls robots to mount the cable onto a C-type clip and a U-type clip. Because of structural differences, the C-type clip expects a larger pushing force during insertion. (b) Multi-modal perception during cable stretching. Upper: camera observations. Lower: visual-tactile images on robot fingertip. The green arrows in ViTac images indicate force vectors, demonstrating a linear force applied along the grasped cable during the stretching process.}
    \label{fig:stretch_process}
    \vspace{-1.8em} 
\end{figure}



On the other hand, learning from demonstrations (LfD) offers an alternative that mitigates the need for prompting examples.
LfD can automatically extract example task plans from demonstrations, eliminating the need for human experts to explicitly construct these plans.
The key questions in the LfD approach for sequential manipulation planning are: 
i) how to segment demonstrations into reusable task plans consisting of skill sequences,
and ii) how to ground the associated skills, which involves determining the necessary task-agnostic information to make skills executable.
Since foundation models are pre-trained on large-scale internet data, it is natural to utilize visual demonstrations~\cite{wake2023gpt} or human play videos~\cite{Wang2023MimicPlay} to address these challenges.

However, visual information alone may be insufficient for perceiving and expressing contact-rich manipulations where the movement of objects is barely observable, yet changes in force contact are significant.
For example, in the cable mounting process shown in Fig.~\ref{fig:stretch_process}, humans intuitively stretch the cable to create tension, thereby easing the subsequent insertion.
As illustrated in Fig.~\ref{fig:stretch_process} (b), while camera-based observations (upper) struggle to capture any movement, images from visual-tactile sensors (lower) clearly reveal a consistent pattern of pulling force.
Even when the stretching step is identified, visual observation alone struggles to define an appropriate success condition for this step.
In such scenarios, force and tactile information play pivotal roles in detecting and understanding such ``invisible" events.
This example leads to a key insight that motivates our work: tactile information is important for both task segmentation, which requires multi-modal semantic reasoning to successfully convert observed long-horizon tasks into a sequence of actions associated with executable robot skills, as well as skill grounding, which extracts necessary information for chaining and executing the skills, such as task specifications and pre/post-conditions.

In this paper, we introduce \textbf{a novel in-context learning framework} that enables pre-trained LLMs to learn task planning from multi-modal demonstration data, with an emphasis on utilizing both visual and tactile information for reasoning about demonstrations and grounding skill conditions.
One single demonstration for each task is provided through teleoperation using a haptic device, where tactile sensing from ViTac sensors and force/torque (F/T) signals are collected alongside camera video recordings.
Each modality is then integrated in a bootstrapped manner into the reasoning process of an LLM analyzer to segment the demonstration into skill sequences, and establish reliable transition conditions between skills.
Subsequently, the resulting sequence and grounded skills serve as an example task plan for an LLM planner, enabling it to generate new plans for varying task configurations.


We evaluate our approach on two challenging, sequential, and contact-rich manipulation tasks: cable assembly and bottle cap tightening. The effectiveness of the proposed framework is validated by experimental results, demonstrating a high success rate in generalizing to novel task configurations. Ablation studies further highlight i) the significant performance improvements in both reasoning and planning when compared to cases without demonstrations or using only visual demonstrations, and ii) the critical role of grounding tactile-based skill conditions for successful reproduction of learned task plan in varied task configurations. These results underscore the importance of incorporating tactile and F/T information, enabling LLMs to effectively understand and learn complex contact-rich manipulation tasks.

\section{Related Work}~\label{sec: related_work}

\textbf{LLMs as Task Planner}.
LLMs have increasingly been used for symbolic task planning due to their strong semantic understanding and reasoning abilities~\cite{huang2024understanding}, though their generated actions aren’t always executable~\cite{zheng2024evaluating}.
To address this, early studies guided LLM plans using predefined action sets and incorporated pre- and post-conditions for better grounding~\cite{zhou2023generalizable}.
Structured approaches like PDDL~\cite{silver2024generalized, zhou2024isr}, behavior trees~\cite{Zhou2024LLMBT,ao2024behavior}, and task trees~\cite{sakib2023cooking} are now commonly applied to improve plan reliability.
Building on this, we also use PDDL for refining a skill library and reasoning about skill sequences.
Another approach to enhancing LLMs/VLMs for task planning is using demonstration datasets. For instance, PaLM-E employs task-and-motion planners to generate extensive planning examples~\cite{10.5555/3618408.3618748}. Other methods use human demonstrations for symbolic task plans but are limited to trajectory-based actions~\cite{wake2023gpt}. Overall, these efforts rely on large datasets and are generally restricted to tasks with limited contact interactions.

\textbf{Learning Task Planning from Few Demonstrations}. 
Classic methods like Hidden Markov Models and Dynamic Motion Primitives learn task planning from limited demonstrations by segmenting tasks into subtasks and learning structured plans chained by subtasks or primitive actions~\cite{konidaris2012robot, ekvall2008robot, manschitz2014learning, kulic2012incremental, niekum2012learning}. 
However, these approaches often require extensive feature engineering and struggle to generalize to new tasks. While imitation learning with deep neural networks has made vast progress in learning and generalizing multi-stage tasks~\cite{zhu2022bottom,mandlekar2020learning, gupta2019relay, shao2021concept2robot, zhou2024language}, it suffers from the high cost of collecting large demonstration datasets.
Leveraging the multi-modal reasoning capability of pre-trained foundation models, LLM-enabled task planning has improved generalization with fewer demonstrations. For instance, GPT-4-based planners achieve one-shot planning from human demonstrations~\cite{wake2023gpt}, and hierarchical structures have been proposed to maximize knowledge distillation from these demonstrations~\cite{chen2023human}. Other methods use task conditions generated by LLMs to guide generalization during execution~\cite{zhou2023generalizable}. Our work extends this research by focusing on how LLMs can effectively utilize demonstrations for contact-rich manipulation task planning.

\textbf{Multi-Modal Sensory Data and LLMs}. Recent advances in multi-modal LLMs, such as GPT-4V~\cite{openai2024gpt4technicalreport}, have exhibited significant capabilities in scene understanding.
However, the current paradigm predominantly focuses on large-scale pretraining for language-conditioned visual representations, while comparatively few studies have investigated other sensory modalities and their potential applications in robotics.

The GenCHiP framework allows LLMs to reason about motion and force by exposing constraints on contact stiffness and forces in control interface~\cite{10801525}, aiming to automate parameter tuning for contact-rich manipulation tasks. Other efforts enhance LLMs' reasoning with tactile data by fine-tuning models on specialized datasets that combine paired tactile-visual observations with tactile-semantic labels~\cite{fu2024touch}. 
Another approach uses contact microphones as tactile sensors to leverage large-scale audio-visual pretraining, addressing the scarcity of non-visual data in low-data robotic applications~\cite{Mejia2024HearingTA}. 
In contrast, our approach minimizes the need for large-scale multi-modal datasets by bootstrapping LLMs with a single human demonstration, integrating ViTac images, F/T signals, and standard camera videos.


\begin{figure*}[t!]
    \centering
    \includegraphics[width=0.98\textwidth]{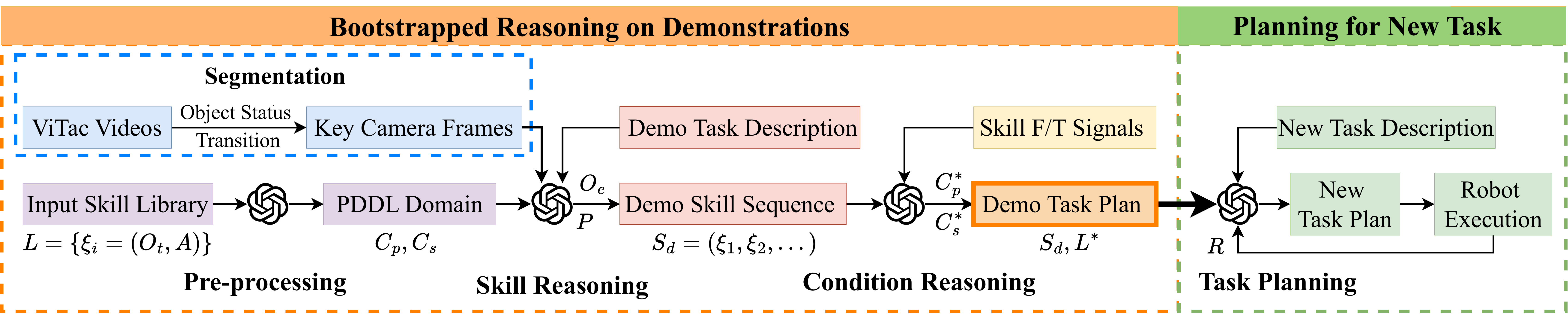}
    \caption{Framework Overview. In bootstrapped reasoning, an LLM analyzer pre-processes the skill library, reasons about skill sequences and success conditions from multi-modal demonstration sequentially. The resulting demo task plan is used as an example for an LLM planner to plan for new tasks.}
    \label{fig: overview}
\end{figure*}

\section{Methodology} ~\label{sec: method}
~\label{sec: skill_library}
\begin{figure*}[t!]
    \centering
    \begin{tikzpicture}
    
    \node [inner sep=0pt] (russell) at (0,0){\includegraphics[width=0.98\textwidth]{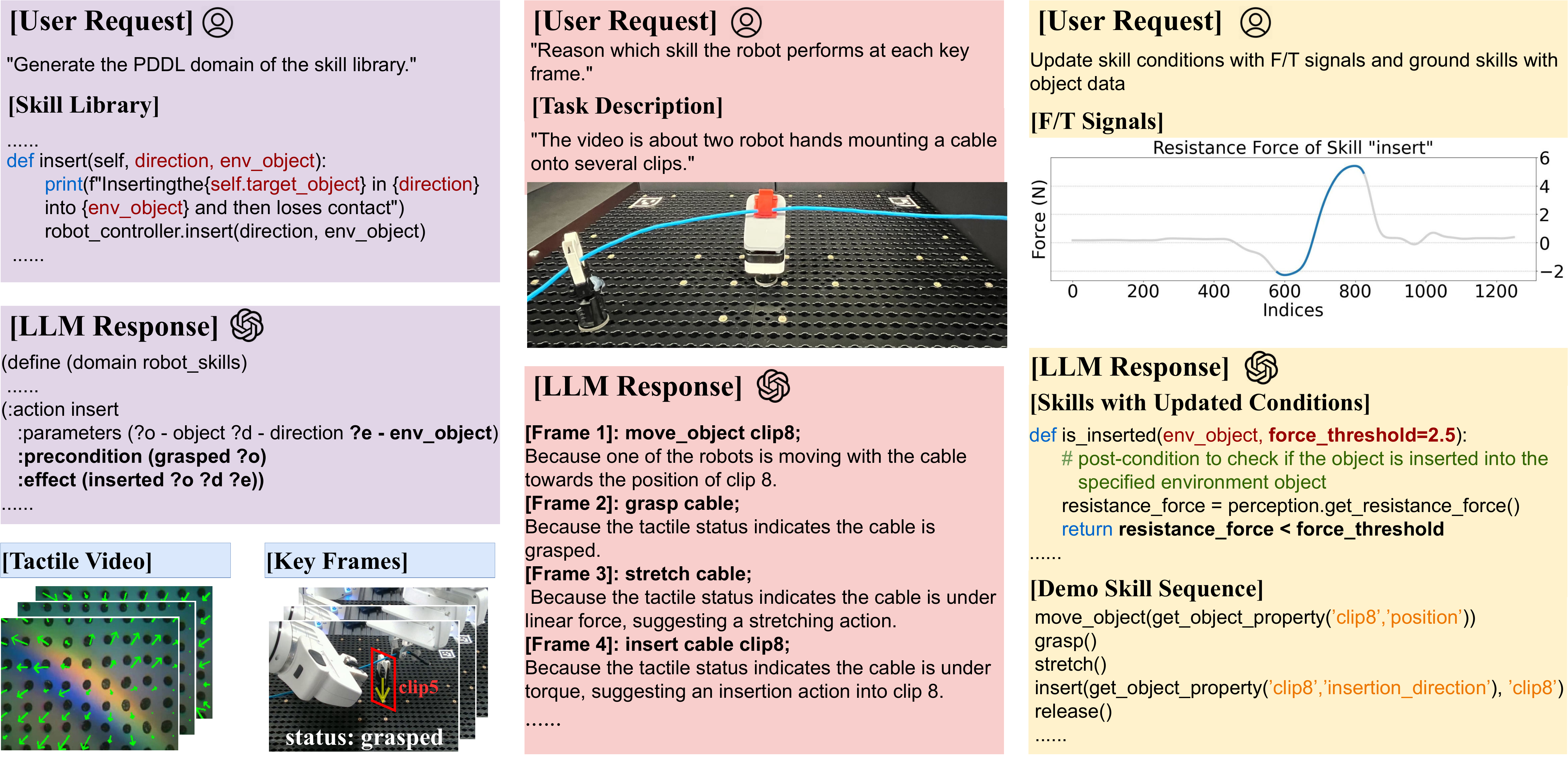}};
    \node at (-5.7,-4.5) {\footnotesize(a) Pre-processing};
    \node at (0,-4.5) {\footnotesize(b) Skill Reasoning};
    \node at (5.7,-4.5) {\footnotesize(c) Condition Reasoning};
    
    \end{tikzpicture}
    \vspace{-1.0em} 
    \caption{Overview of Prompts and Responses in Bootstrapped Reasoning. More prompts can be found on our project website.}
    \label{fig:prompt_overview}
    \vspace{-1.5em} 
\end{figure*}
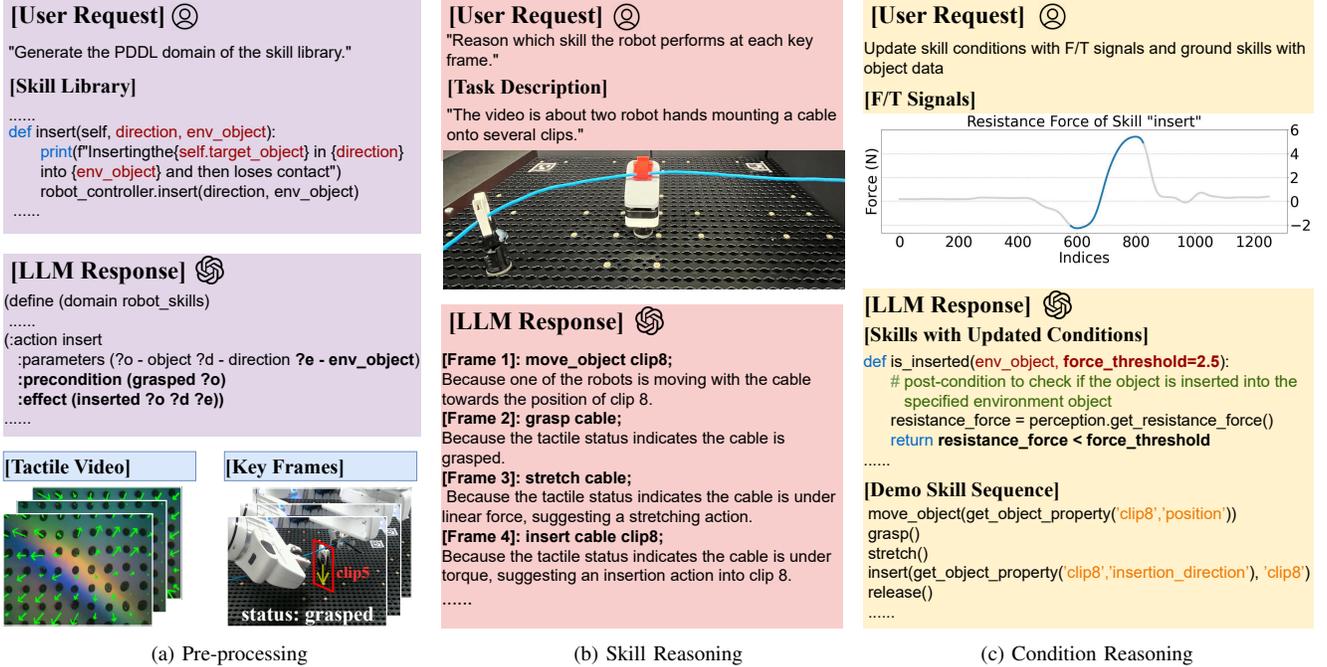

As shown in Fig.~\ref{fig: overview}, our framework first derives a task plan from the demonstration (highlighted in the orange box), and then uses this plan as a reference for planning new, generalized tasks (highlighted in the green box). 
Prior to this, we collect ViTac data from the robot’s fingers, third-person camera images, and F/T measurements at the robot's end-effector during kinesthetic teaching.
In the reasoning stage, in addition to the skill library, we provide the LLM analyzer with each sensory modality in a bootstrapped manner, enabling it to derive from the demonstration a task plan that describes the demonstration as well as a skill library with transition conditions updated.
When a new task is requested, the LLM planner uses the demonstration task plan as an example to generate new task plans, adjusting them interactively based on the robots' execution feedback.

In the following section, we first introduce the formulation of our object-centric skill library. 
Next, we explain each stage of the bootstrapped reasoning pipeline, presenting how each modality is integrated.
Finally, we describe how the generated task plan is utilized for planning on new tasks.

\subsection{Object-Centric Skills} 
In contact-rich manipulation tasks, such as assembly, robot actions are typically highly object-centric. 
That is to say, each robot action is designed to change a certain status of an object, whether by changing its position or applying an external force.
In this work, we focus on two key aspects of object status, namely, positional status (which includes the object's position and orientation) and interaction status (which describes whether the object is grasped, released, or subjected to an external force or torque).

Inspired by this fact, we construct our skill library in an object-centric manner, where each skill is defined to describe a status change of the being-manipulated object (called target object). 
In this way, the formulation of skills remains consistent whether executed by a human or a robot, enabling LLMs to generalize human demonstrations into executable task plans in new task configurations.
Additionally, we utilize the target object status as the bridge between raw sensory data from demonstrations and the grounding of LLM, which will be discussed in detail in Section~\ref{sec: reasoning}.

\textbf{Input Skill Library.} Each skill in our skill library is formulated as a tuple $\xi = (O_{t}, O_{e}, C_{p}, C_{s}, A, R)$.
$O_{t}$ represents the target object which robot directly manipulates, such as a cable or a cap. 
$O_{e}$ represents the contextual object with which the $O_{t}$ interacts under the manipulation of robots, such as a cable clip or a bottle. 
The description and examples of all components in $\xi$ are summarized in Table~\ref{tab: skill_form}.

\begin{table}[ht]
\footnotesize
\caption{Skill Formulation}
\centering
\resizebox{0.48\textwidth}{!}{
\begin{tabular}{p{0.15cm}|p{3.8cm}|p{2.2cm}|p{1.6cm}}
 \toprule
&Description &Example &PDDL   \\
 \hline
$O_{t}$ &\textbf{Target object} that the robot directly manipulates. 
& cable &\ftnotett{parameters} \\
 \hline
$O_{e}$ &\textbf{Contextual object} with which the $O_{t}$ interacts. & clip &\ftnotett{parameters} \\
\hline
$A$ &Executed \textbf{Action} by low-level robot controller.
& insertion & \ftnotett{action} \\
 \hline
$C_{p}$ &\textbf{Pre-condition} to be met for the skill to start execution. 
& cable positioned near a clip  &\ftnotett{precondition} \\
 \hline
$C_{s}$ &\textbf{Success condition} that defines when the skill is finished. 
& cable inserted into a clip &\ftnotett{effect} \\
 \hline
$R$ & \textbf{Return} of a skill execution. & success (when $C_{s}$ is satisfied) or error & N/A\\
 \hline
$P$ &\textbf{Skill parameters}, such as force thresholds or positions. 
&position of the clip & \ftnotett{parameters}\\

 \bottomrule
\end{tabular}} \label{tab: skill_form}
\vspace{-1.3em}
\end{table}
Our skill library $L=\{\xi_i\}$, serving as input to the LLM analyzer, is defined in the form of executable code scripts.
Initially, we include object-agnostic skills, such as \smalltt{move}, in a general {\smalltt{ObjectSkillLibrary}}. 
These skills can be applied to various objects and serve as the foundation for object-specific libraries, which contain affordances of specific objects. 
For example, the \smalltt{CableSkillLibrary} includes a ``cable-centric" \smalltt{insertion} skill (see Fig.~\ref{fig:prompt_overview}(a)), which handles inserting a cable into other objects, such as clips.
The definition of each skill provides explicitly $O_{t}$ and calls the low-level \smalltt{robot\_controller} interface to perform an action $A$.
The \smalltt{robot\_controller} controls the robot to move or apply force/torque via an adaptive impedance controller~\cite{johannsmeier2019framework}.
In the following procedures, we will guide the LLM to complete the remaining components in skill $\xi$ step by step.


\textbf{Pre-processing Translation.} While the original formatting of code scripts allows skills to be executed by robots, it lacks the logic to support effective reasoning, especially conditions for skill transitions.
Inspired by prior works~\cite{silver2024generalized, zhou2024isr}, we ask the LLM analyzer to translate the input skill library into a PDDL domain as a preprocessing step.
As a standard language to describe planning problems, PDDL provides a structured syntax to represent rules like actions, objects, and conditions in its domain knowledge, which closely aligns with the composition of our input skill library.
A comparison between our skill library and the corresponding PDDL domain components is summarized in Table~\ref{tab: skill_form}.
Through translation, we aim to leverage PDDL's standard format to enhance LLMs' understanding of skills and the ability to plan for long-horizon tasks. 
As is shown in Fig.~\ref{fig:prompt_overview}(a), translation into PDDL encourages the LLM analyzer to automatically complete the transition conditions $C_{p}$ (\smalltt{precondition} in PDDL) and $C_{s}$ (\smalltt{effect} in PDDL).

\subsection{Bootstrapped Reasoning of Demonstration}\label{sec: reasoning}

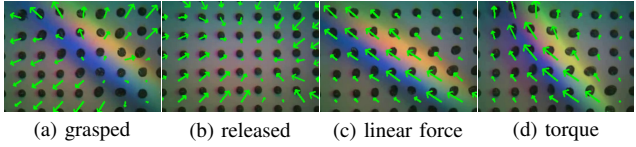
\begin{figure}[t!]
\centering
\begin{tikzpicture}

    \node at (-.5,0) {\includegraphics[width=0.23\columnwidth, trim={1.8cm 1.8cm 1.8cm 1.8cm},clip]{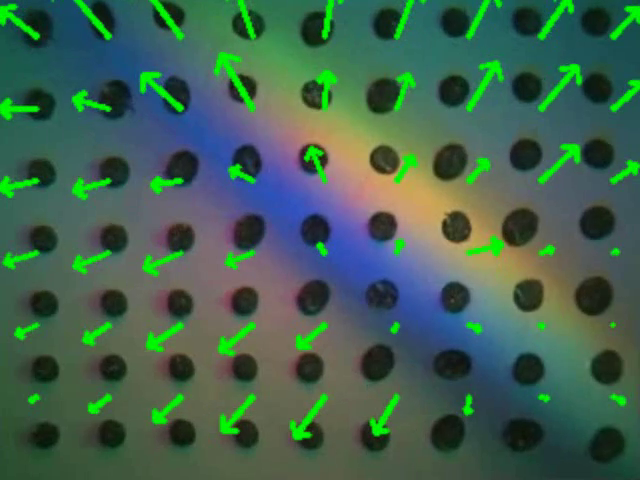}};
    \node at (-.5,-1.0) {\footnotesize(a) grasped};

    \node at (1.6,0) {\includegraphics[width=0.23\columnwidth, trim={1.8cm 1.8cm 1.8cm 1.8cm},clip]{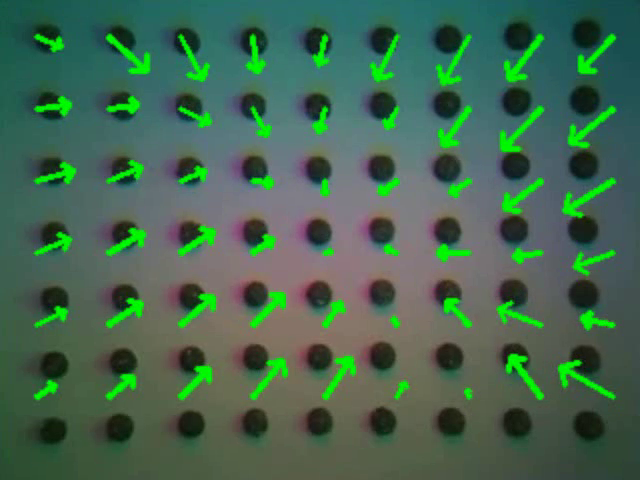}};
    \node at (1.6,-1.0) {\footnotesize(b) released};

    \node at (3.70,0) {\includegraphics[width=0.23\columnwidth, trim={1.8cm 1.8cm 1.8cm 1.8cm},clip]{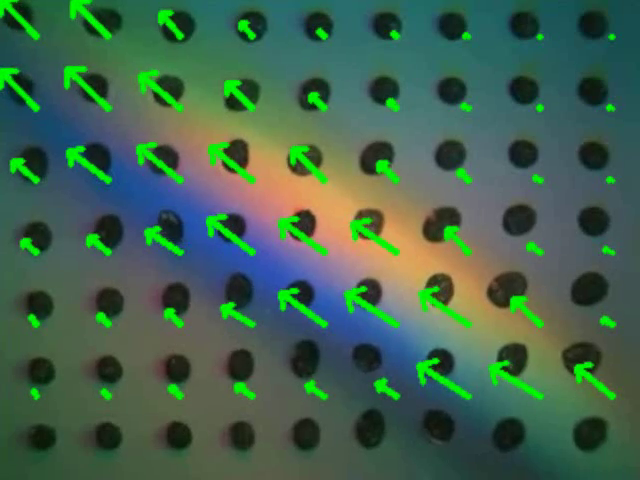}};
    \node at (3.7,-1.0) {\footnotesize(c) linear force};

    \node at (5.80,0) {\includegraphics[width=0.23\columnwidth, trim={1.8cm 1.8cm 1.8cm 1.8cm},clip]{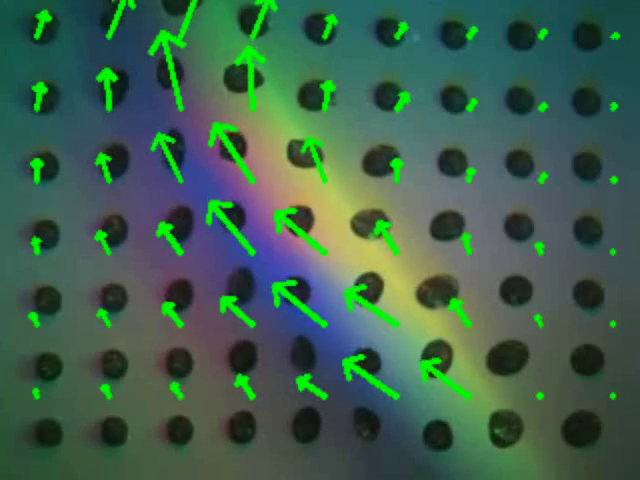}};
    \node at (5.80,-1.0) {\footnotesize(d) torque};
\end{tikzpicture}
\vspace{-1.0em}
\caption{Tactile Signal Patterns and Corresponding Object Statuses: (a) Sourcing pattern, referring to ``grasped" status; (b) Sinking, referring to ``released" status; (c) Uniform Flow, referring to ``under a linear force" status; (d) Twisted Flow, referring to ``under torque" status.}
\label{fig:force signal patterns}
\vspace{-1.5em}
\end{figure}

Given the large volume and multi-modal nature of our demonstration data, it can be challenging for LLMs to interpret all modalities simultaneously. 
To address this problem, we adopt a bootstrapped approach for in-context learning of demonstration, where each modality is introduced sequentially to assist the LLM in different stages of reasoning.
Tactile information is utilized to segment events from visual perception and identify object statuses, which then enables LLMs to comprehend the entire demonstration and infer the corresponding skill sequence $S_d = (\xi_1, \xi_2, ...)$. 
Afterwards, F/T signals are leveraged to ground and refine the transition conditions between skills ensuring that the learned skill sequence is executable and can generalize to new task scenarios by re-planning.

\textbf{Segmentation with Object Status.} 
To facilitate skill reasoning from demonstration data, we first segment the entire demonstration into events where object status changes.
We rely on tactile status for this segmentation, as tactile information on robot fingers directly precepts the manipulation on target object $O_{t}$, thus capturing details that are often missed by third-person cameras.

Fig. \ref{fig:force signal patterns} illustrates four typical patterns of Vi-Tac images, each reflecting different interaction status of the target object: 
a) \textit{Sourcing}: Vectors point outward, indicating a source where force radiates outward. This pattern is usually observed when an object is ``grasped".
b) \textit{Sinking}: Vectors point inward, showing a sink where force converges inward, usually when a grasped object is ``released".
c) \textit{Uniform Flow}: Vectors are parallel and evenly spaced, representing a consistent force in one direction. This pattern happens when an object is ``under a linear force", such as being pushed or pulled.
d) \textit{Twisted Flow}: Vectors twist gently, suggesting rotational movement around a central point. This pattern happens when an object is ``under torque", such as being rotated or screwed.

To identify the above tactile patterns in demonstrations, we fine-tune the video classifier \textit{TimeSformer}~\cite{gberta_2021_ICML} with a dataset of labeled tactile videos.
Applying this classifier to the complete demonstration then segments it into events when new interaction happens.
Object status alongside the timestamps of these events (referred to as key timestamps) are extracted for subsequent reasoning.


\textbf{Reasoning Skill Sequence}
After segmentation, we provide the LLM analyzer with camera images taken at key timestamps (referred to as key frames) for skill reasoning. 
As is shown in Fig.~\ref{fig:prompt_overview} (a), each key frame is annotated with objects as well as the the corresponding status of $O_{t}$.
Additionally, we include in the prompt a simple description of the task (e.g., ``two robots are mounting a cable onto several clips") and of the objects (e.g., ``the blue curve in the view is the cable"). 
The LLM is then asked to reason which skill from the PDDL domain the demonstrator has performed at each key timestamp.




Fig.~\ref{fig:prompt_overview}(b) presents an example of the skill sequence reasoned by the LLM. 
We observe that the LLM adaptively utilizes camera images (for \smalltt{move\_object}) and object status (for \smalltt{grasp}, \smalltt{stretch}, and \smalltt{insert}) to infer the corresponding skills. 
In addition, the LLM also fills in the contextual object $O_e$ (\smalltt{env\_object}) for each skill in its reasoning.

\textbf{Reasoning Skill Conditions}
The final step in converting the skill sequence into an executable task plan is to reason about success conditions for each skill. 
We leverage the \smalltt{precondition} and \smalltt{effect} in the previously translated PDDL domain.
The LLM is requested to implement these conditions and integrate them back into the skill library. 
Additionally, we provide the LLM with interfaces to access perception information about robot pose, grasping status, and F/T signals. 
Since most of the other information is either binary or straightforward when used to form conditions (e.g. whether an object is grasped or a position is reached), we focus especially on F/T conditions which are highly variable and crucial for contact-rich manipulations.

The raw six-dimensional F/T signals are complex for the LLM to interpret directly.
To address this without sacrificing generality, we assume that the task is performed in a static environment where interactions with the object occur exclusively through the robot. 
In this context, the most relevant F/T information pertains to the force or torque opposing the robot's actions, as they provide direct feedback on the resistance encountered during manipulation.
Based on this observation, we reduce our F/T perception interface to include only resistance force $f_r$ and torque $\tau_r$.

For each skill, we first ask the LLM to generate an initial success condition function, in which it determines which signal the condition should be based on (e.g. \smalltt{resistance\_torque} is used to form the \smalltt{is\_tightened} condition). 
We then provide a plot of the selected signal and prompt the LLM to update success condition functions accordingly.
An example of the resulting function \smalltt{is\_inserted} for the task of mounting cable to clips is shown in Fig.~\ref{fig:prompt_overview}(c). 
The LLM defines the success condition for insertion as the resistance force falling below a certain threshold, indicating that the cable has been securely inserted. 
We present later in Section~\ref{subsec: reason_eval} that numerical thresholds are particularly refined with the F/T signals.
% ...

% ...


\subsection{Planning in New Scenarios}
From the above bootstrapping reasoning, the LLM analyzer has extracted the skill sequences  corresponding to human demonstrations, and has extended the original skill library with appropriate success conditions.
These outputs are then combined as an demonstration task plan, which will be used as an example for task planning on a new task.
As shown in Fig.~\ref{fig: overview}, in the planning request to an LLM planner, we include the demonstration task plan as well as an image and description of the new task scene.
To make the plan more dynamic and flexible, we also use the LLM planner to monitor the execution process.
After execution of each skill, we feed its return $R$ back to the LLM planner, which then decides whether the plan should be continued or adjusted.

\section{Experiments}
In this section, we present real-world experiments to evaluate the effectiveness of our demonstration reasoning pipeline and planning results for new tasks.
This evaluation is conducted through ablation study: by disabling or replacing a certain parts in our framework, we design the following control groups.
\begin{itemize}
    \item[A.] Transition Frames Without Object Status: Key frames in our demonstration reasoning pipeline are replaced by frames at key timestamps but without status annotation.
    \item[B.] Uniform Sampled Frames Without Object Status: Similar to (a), but frames are sampled uniformly from video, again without status annotation. 
    \item[C.] Conditions Without F/T Signals: Force/torque signals are excluded from the demonstration reasoning pipeline, so the success conditions remain as initially generated by the LLM without any updates.
    \item[D.] Without Demonstrations: No demonstration data is provided and the LLM generates the plan solely based on its prior knowledge. 
\end{itemize}

\subsection{Experimental Setup}


\begin{figure}[!t]
    \centering
    \begin{tikzpicture}
    \node[inner sep=0pt] (russell) at (0.0,0.0)
    {\includegraphics[width=0.48\textwidth]{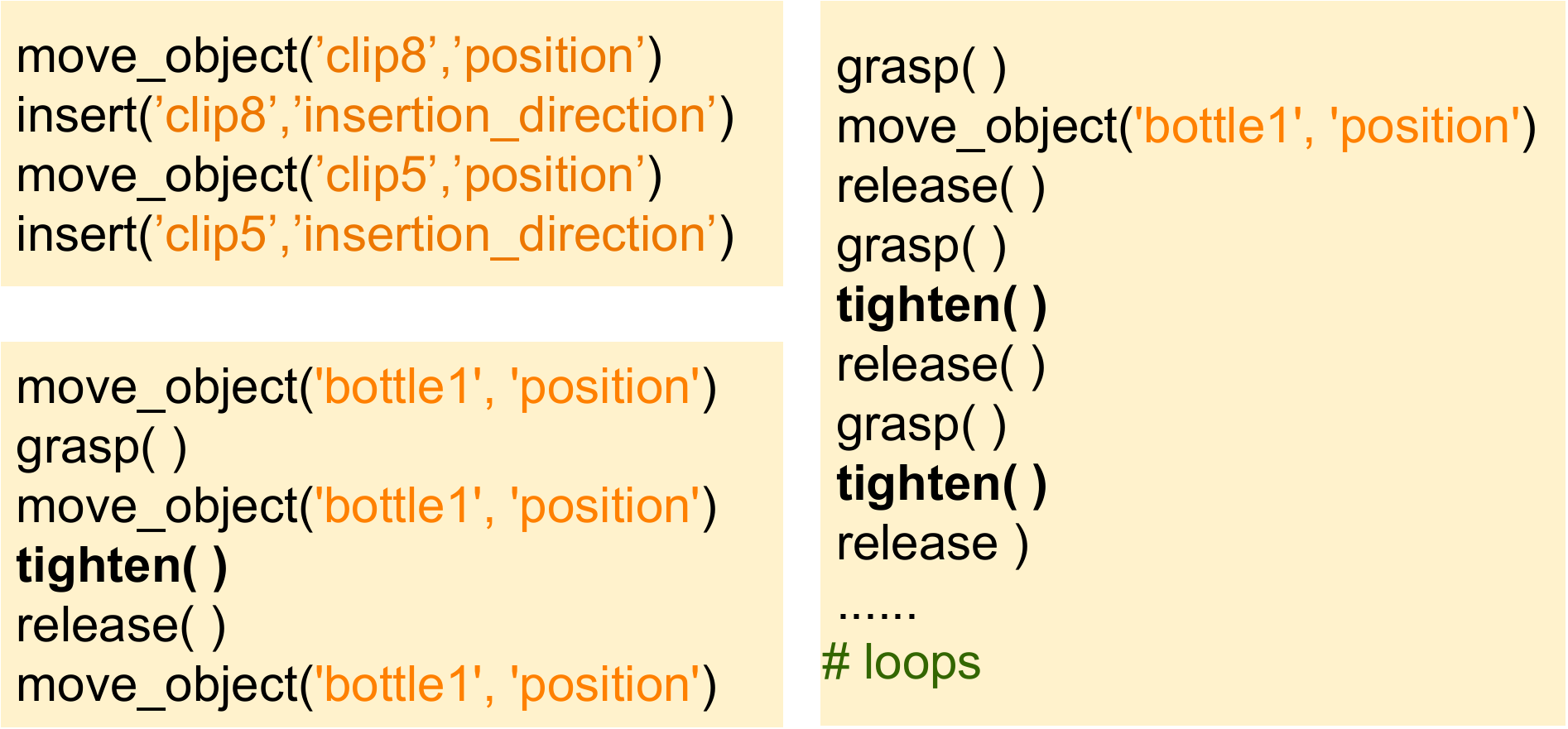}};
    \node at (-2.2,0.25) {\footnotesize(a) Control Group A and B};
    \node at (-2.2,-2.2) {\footnotesize(b) Control Group A};
    \node at (2.2,-2.2) {\footnotesize(c) Our Pipeline};
    \end{tikzpicture}
    \vspace{-2.0em} 
    \caption{Skill Sequences Reasoned from Demonstrations. (a) Sequence reasoned for cable mounting by control groups A and B. The demo skill sequence in Fig.~\ref{fig:prompt_overview}(c) shows the result by our pipeline. (b) and (c) Sequence reasoned for cap tightening. The resulting sequence of group B is similar to group A but with more move\_object steps.}
    \label{fig: reason_comparison}
    \vspace{-1.7em} 
\end{figure}

\begin{figure*}[!t]
    \centering
    \begin{tikzpicture}
        \node[anchor=north west] (img1) at (-2.25, 0) {\includegraphics[height=0.13\textwidth]{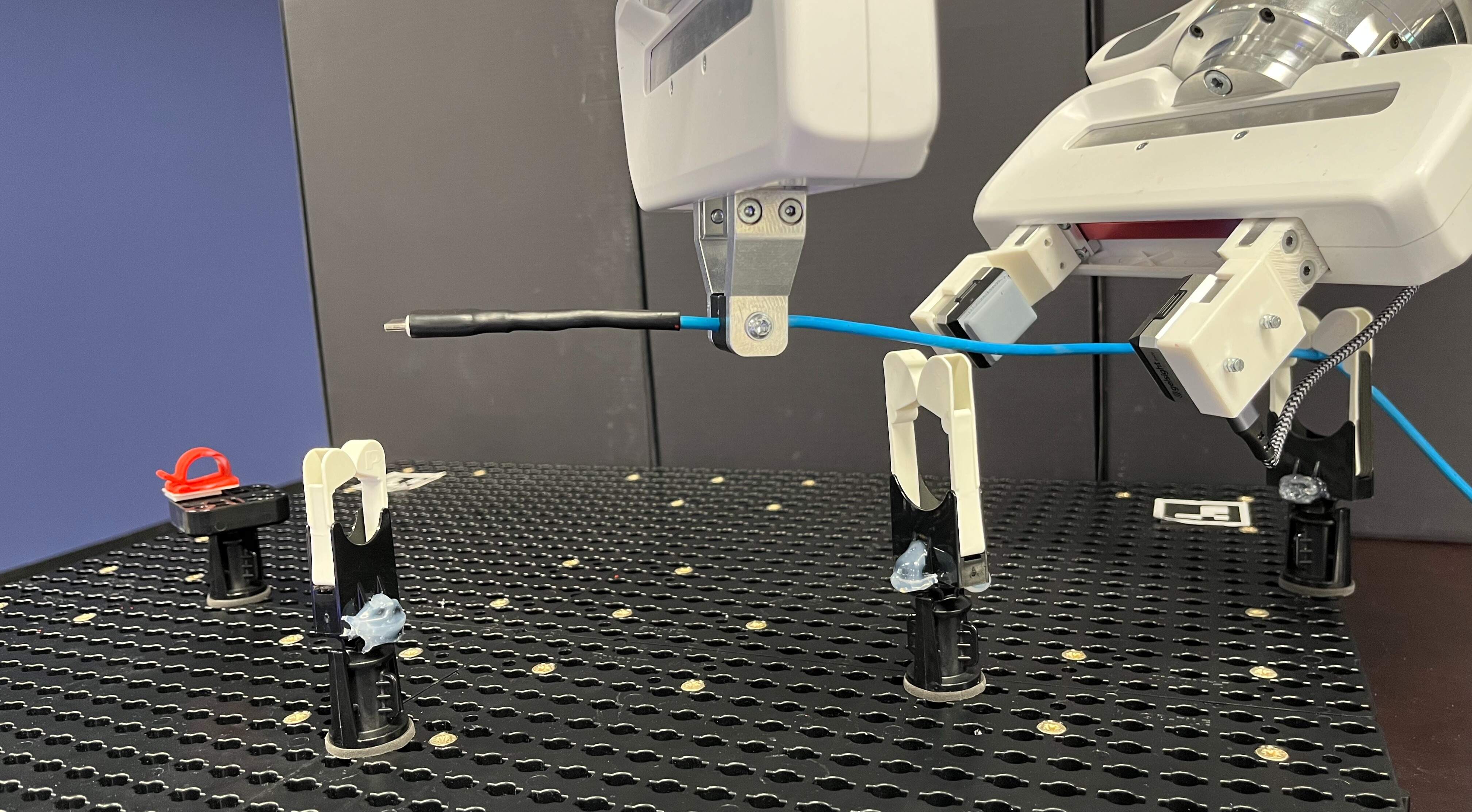}};
        \node[anchor=north west] (img2) at (2.3, 0) {\includegraphics[height=0.13\textwidth]{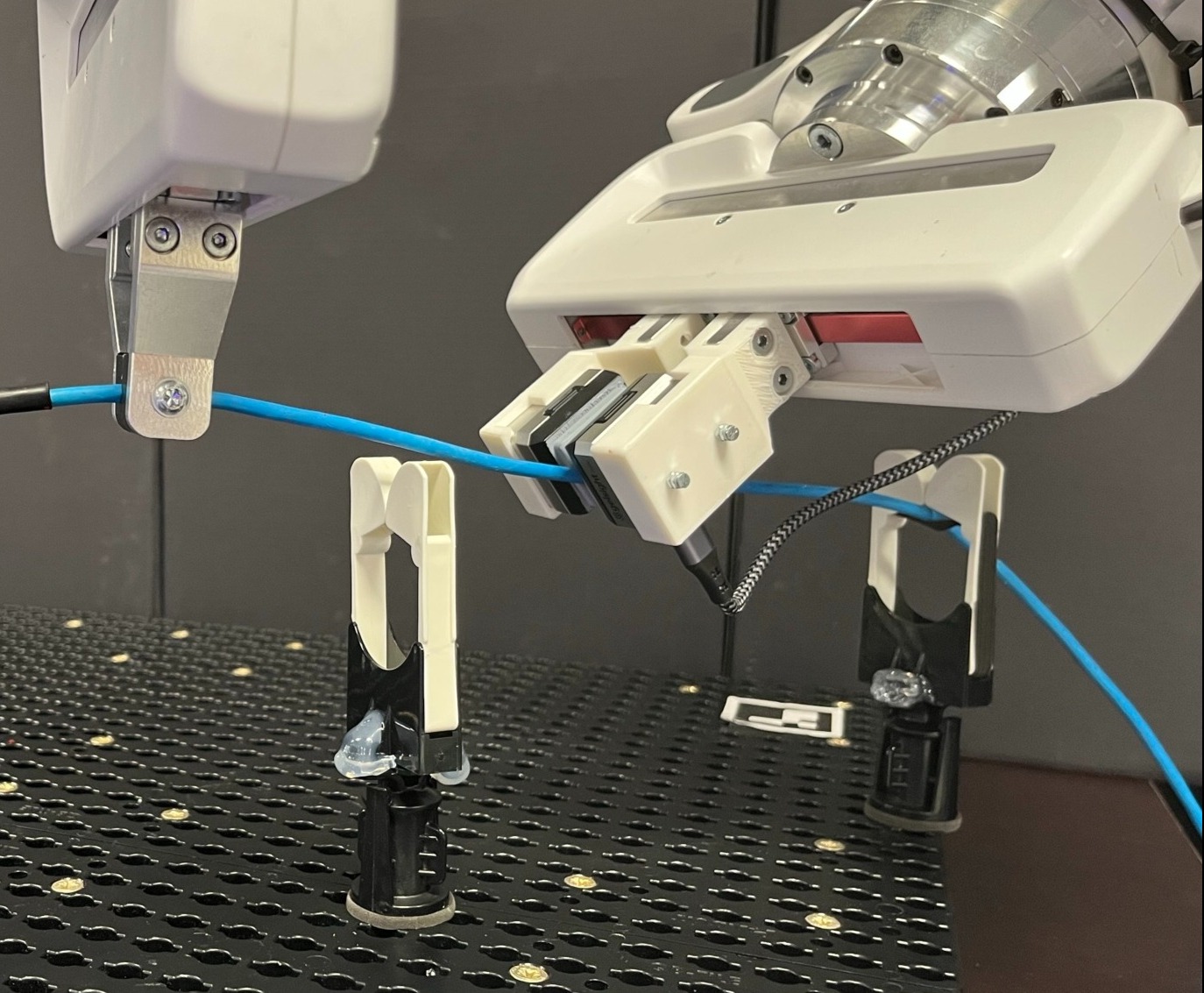}};
        \node[anchor=north west] (img3) at (5.5, 0) {\includegraphics[height=0.13\textwidth]{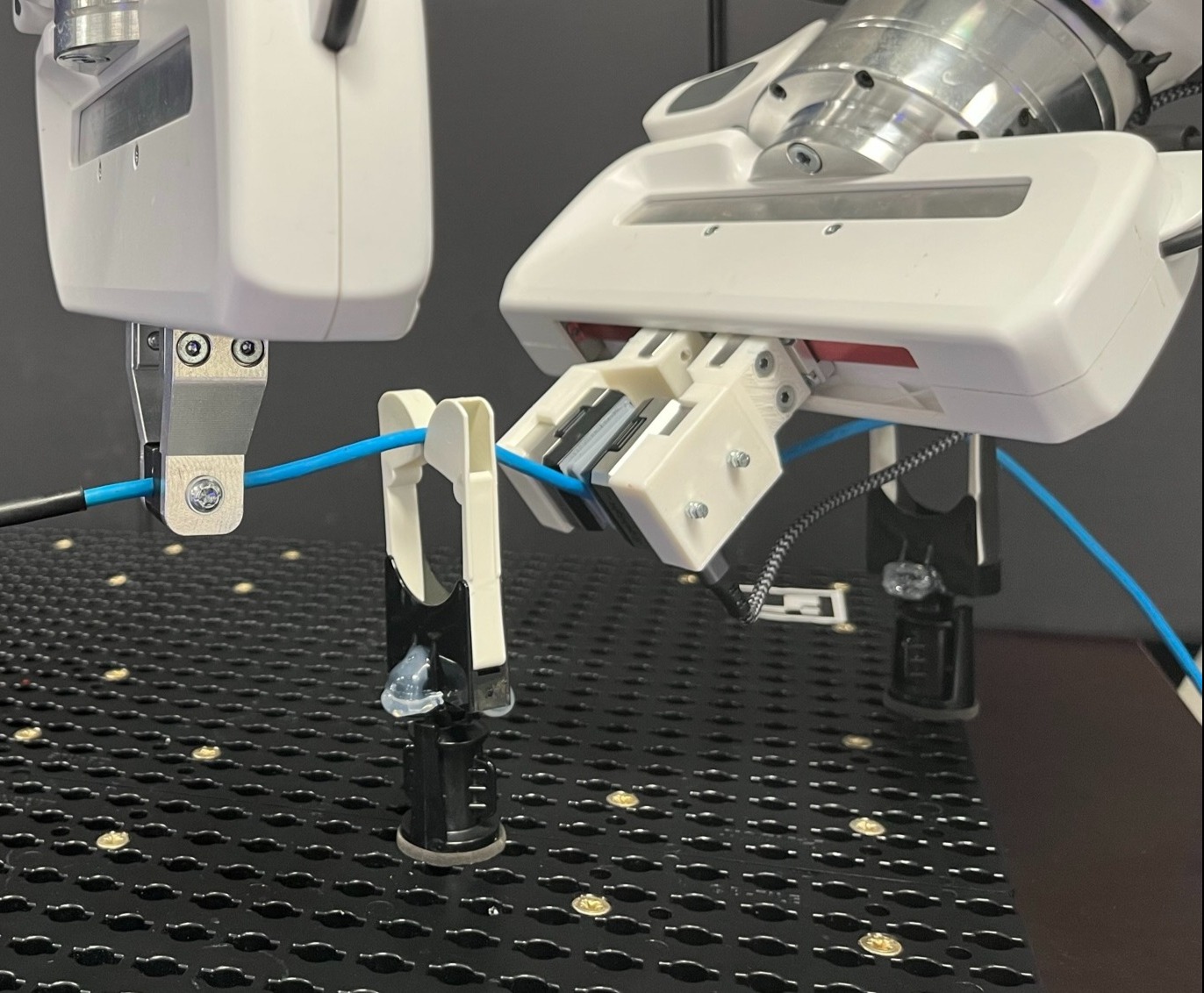}};
        \node[anchor=north west] (img4) at (8.70, 0) {\includegraphics[height=0.13\textwidth]{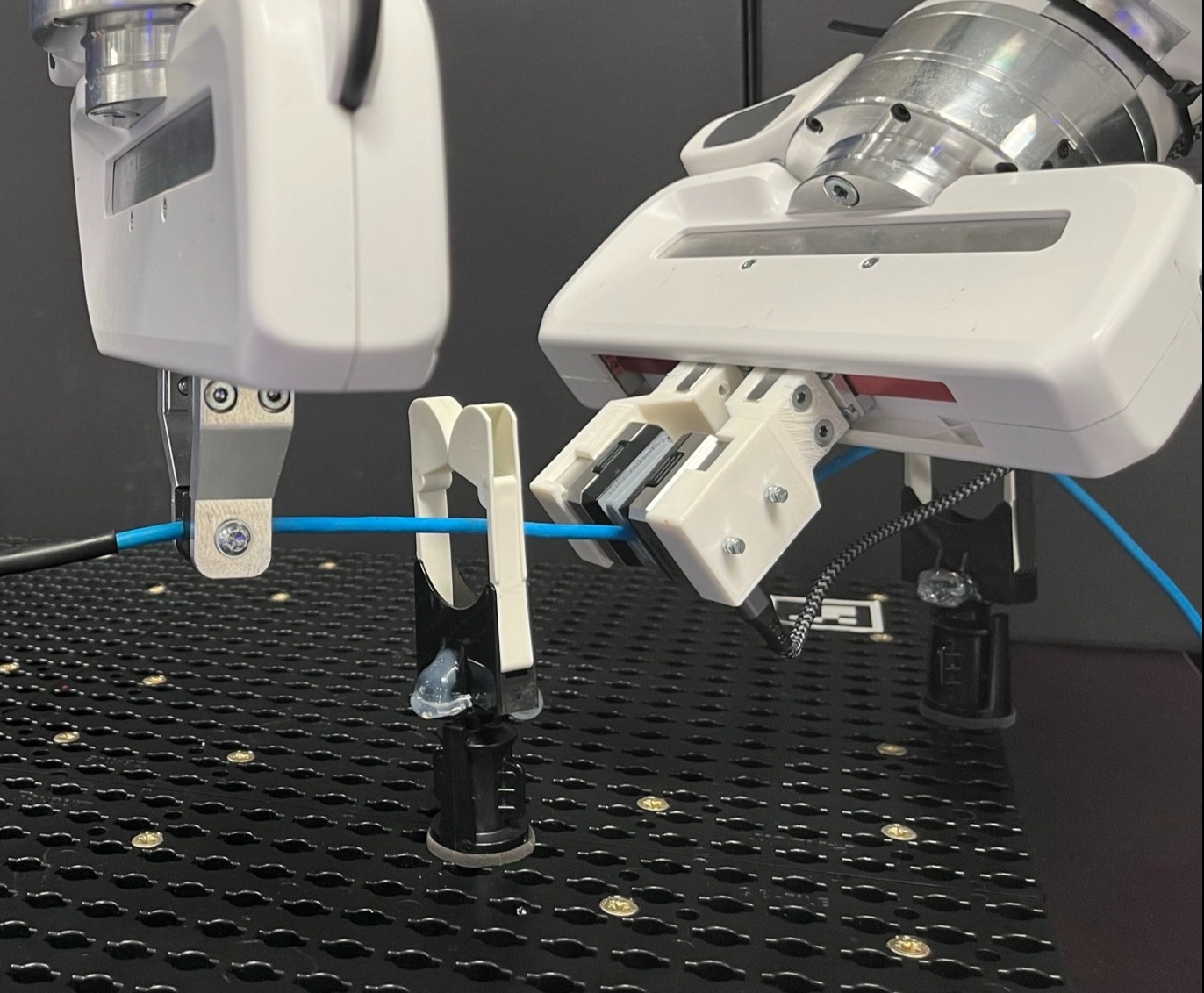}};
        \node[anchor=north west] (img5) at (11.9, 0) {\includegraphics[height=0.13\textwidth]{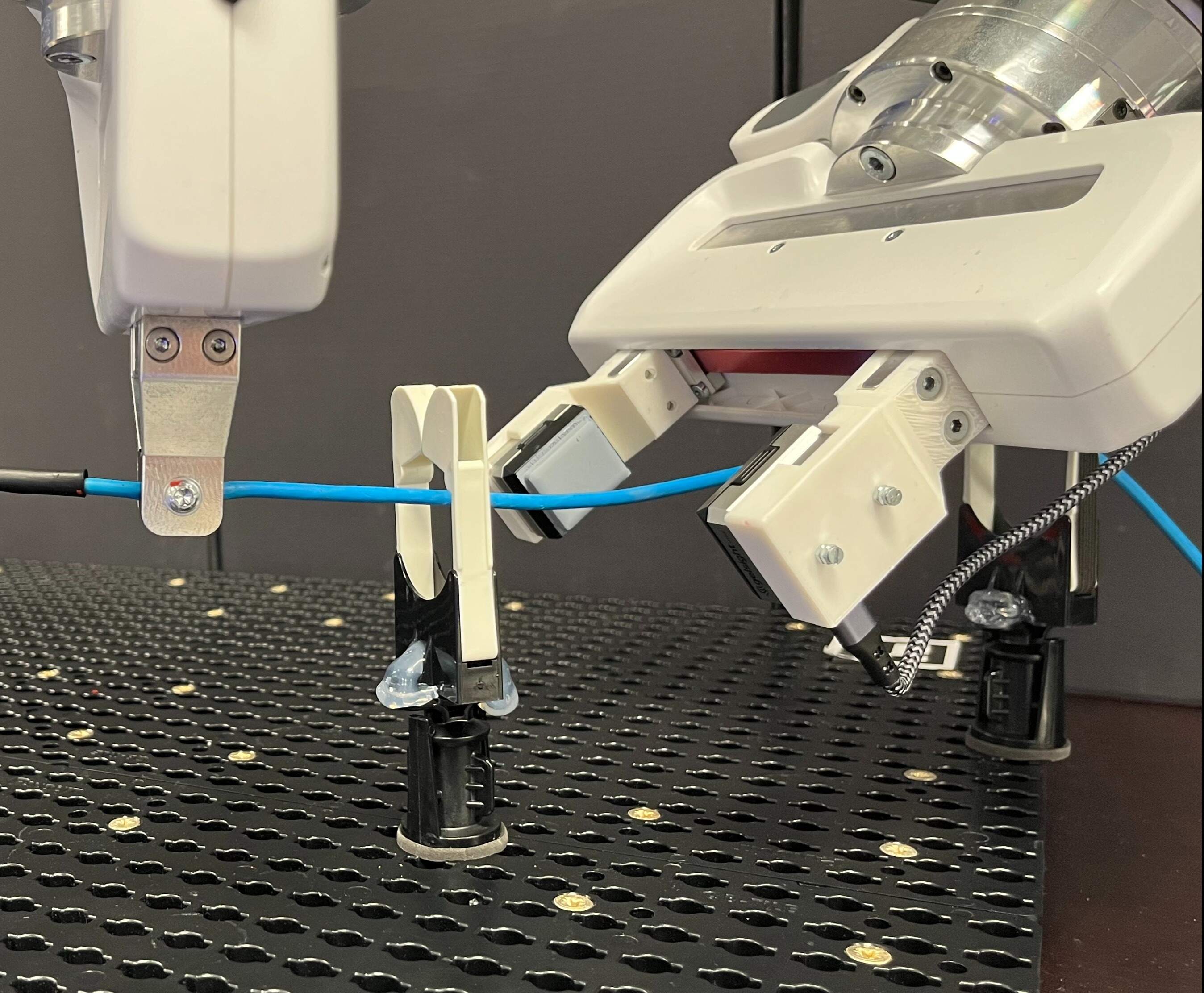}};
        
        \node[anchor=north west] (img6) at (-2.25, -3) {\includegraphics[height=0.13\textwidth]{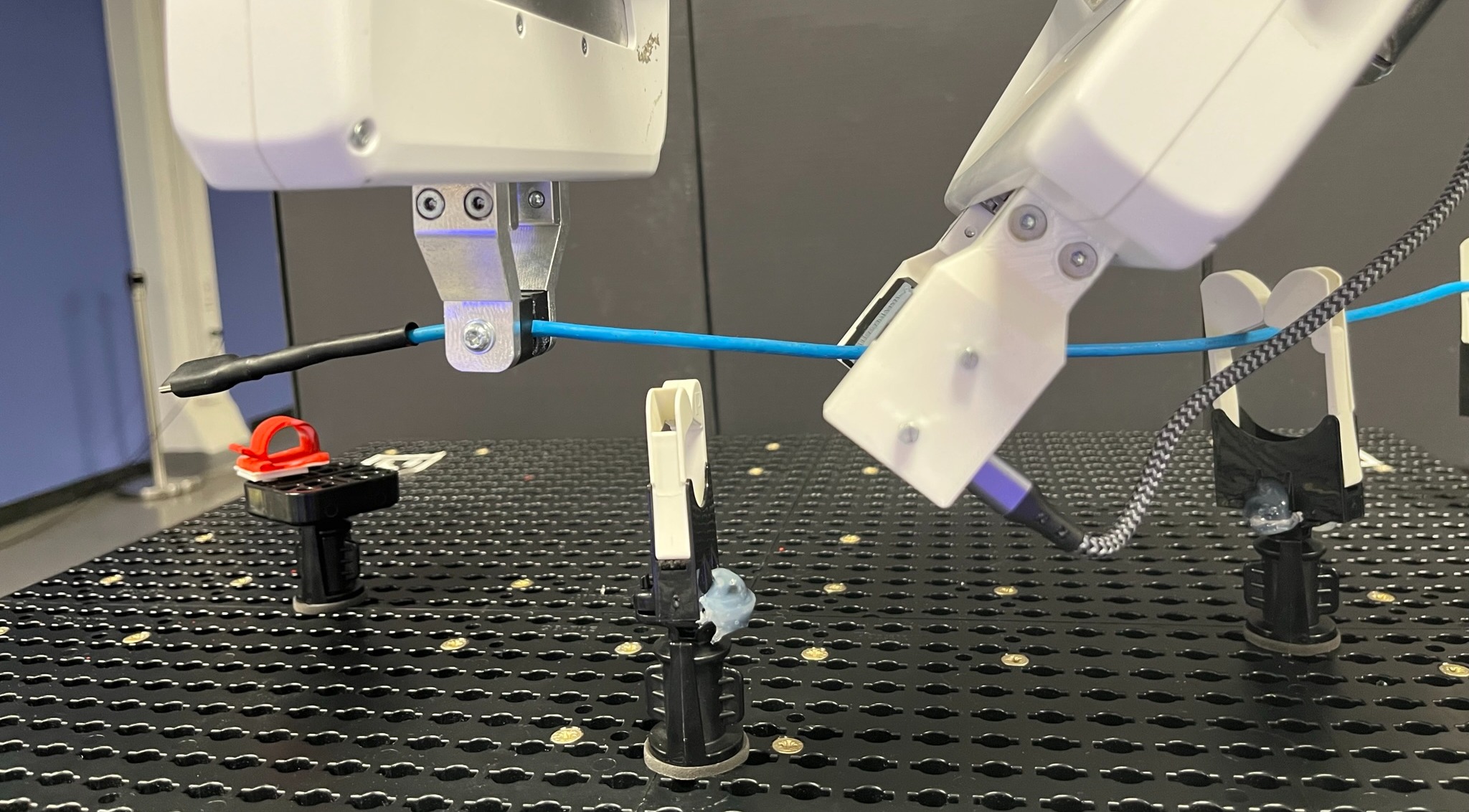}};
        \node[anchor=north west] (img7) at (2.0, -3) {\includegraphics[height=0.13\textwidth, trim={0cm 0cm 0cm 0},clip]{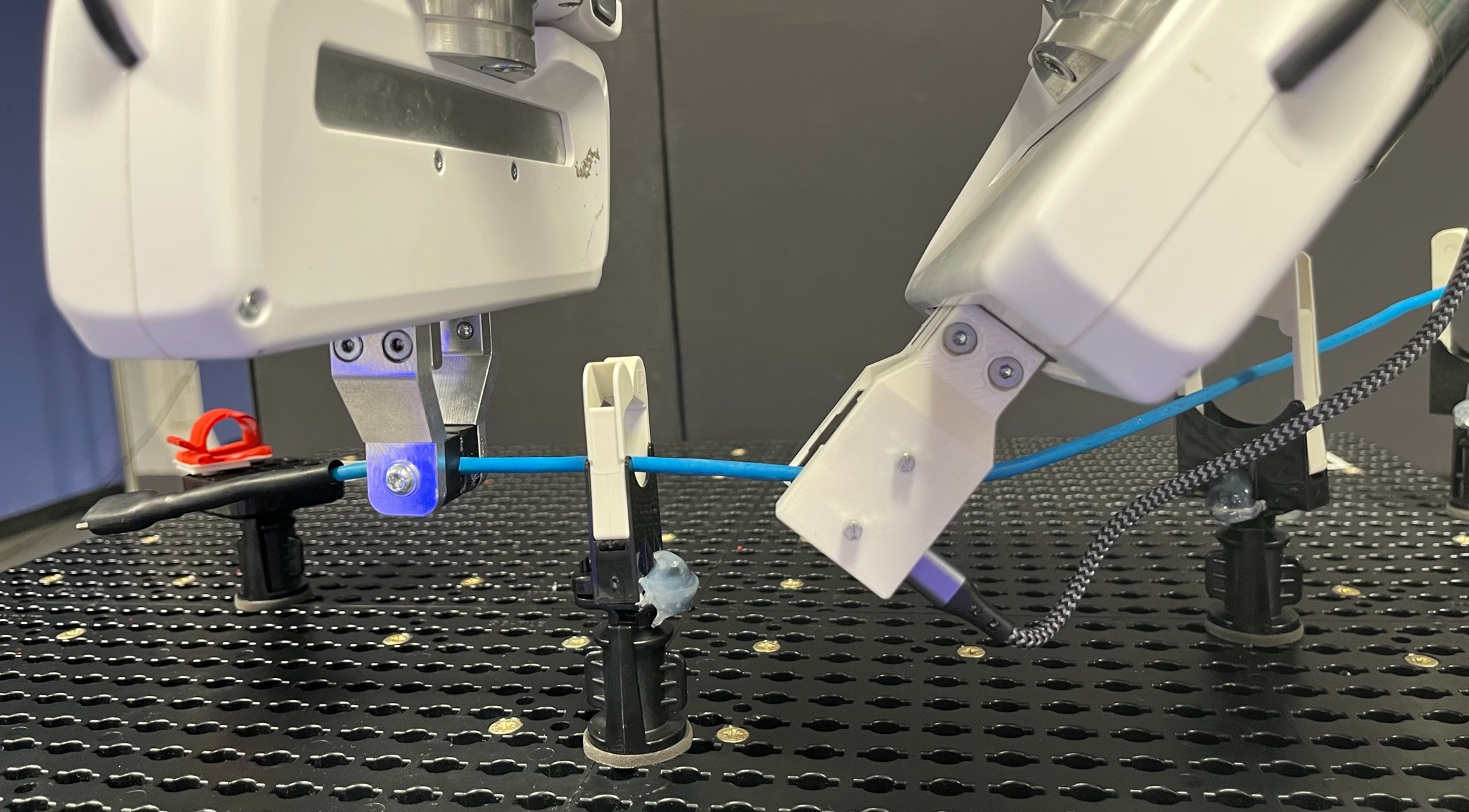}};
        \node[anchor=north west] (img8) at (6.25, -3) {\includegraphics[height=0.13\textwidth, trim={0cm 0cm 0cm 0},clip]{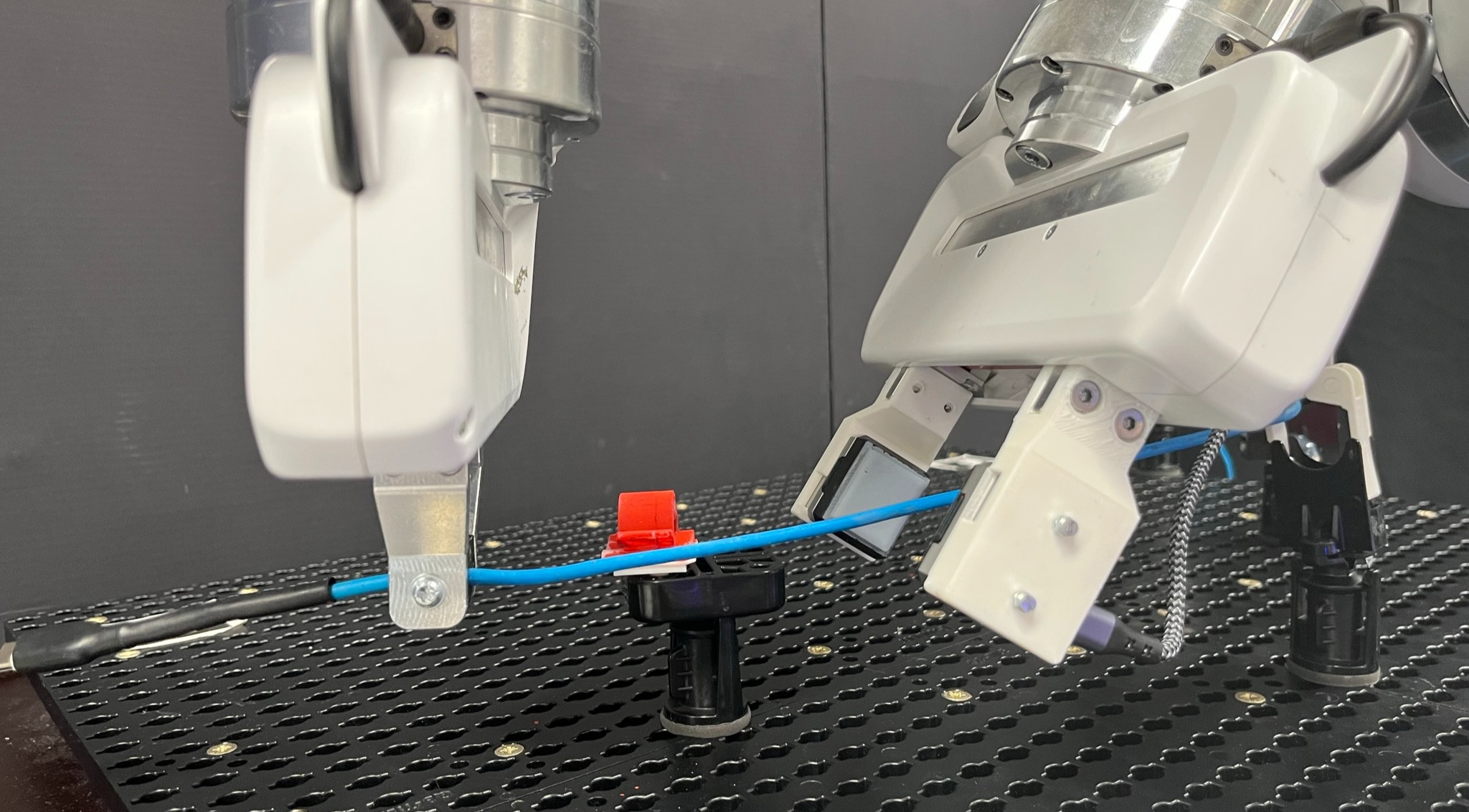}};
        \node[anchor=north west] (img9) at (10.5, -3) {\includegraphics[height=0.13\textwidth, trim={0cm 0cm 0cm 0},clip]{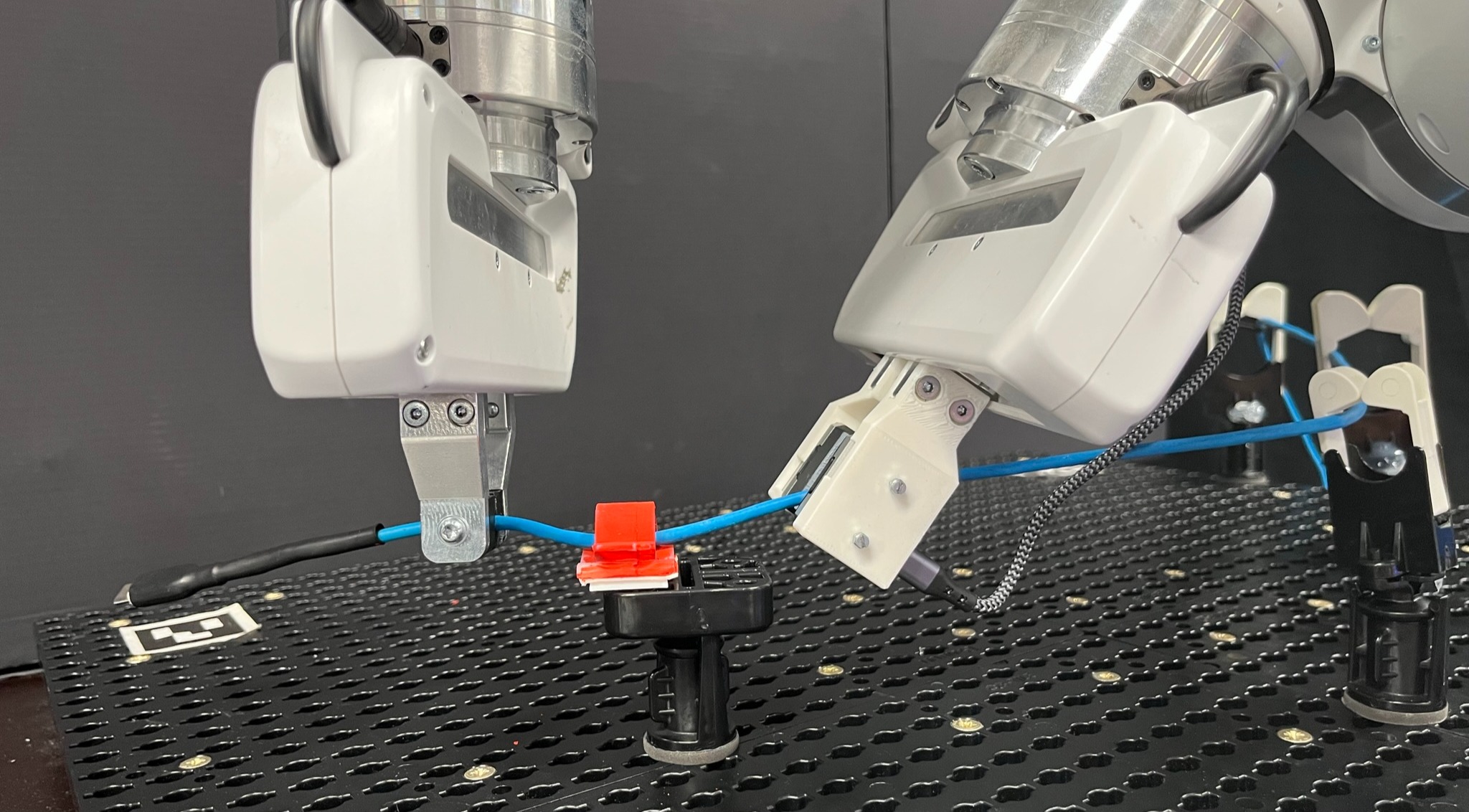}};
        
        \node[anchor=center] (cap1) at (-0.0, -2.70) {move\_object (clip5)};
        \node[anchor=center] (cap2) at (2.9, -2.76) {grasp};
        \node[anchor=center] (cap3) at (4.70, -2.7) {stretch};
        \node[anchor=center] (cap4) at (8.7, -2.7) {insert (clip5)};
        \node[anchor=center] (cap5) at (13.55, -2.7) {release};

        \draw[->, thick] (cap1.east) -- (2.43, -2.70);
        \draw[->, thick] (3.40, -2.7) -- (cap3.west);
        \draw[->, thick] (cap3.east) -- (cap4.west);
        \draw[->, thick] (cap4.east) -- (cap5.west);
        \draw[->, thick] (cap5.east) -- (14.8, -2.70);

        \node (cap6) at (0.9, -5.7) {stretch};
        \node (cap7) at (4.2, -5.7) {insert (clip6)};
        \node (cap8) at (8.5, -5.7) {grasp};
        \node (cap9) at (12.5, -5.7) {insert (clip8)};

        \node (cap10) at (-0.9, -5.7) {…};
        \node (cap11) at (6.6, -5.7) {…};
        \node (cap12) at (14.5, -5.7) {…};

        \draw[->, thick] (-2.1,-5.7) -- (cap10.west);
        \draw[->, thick] (cap10.east) -- (cap6.west);     
        \draw[->, thick] (cap6.east) -- (cap7.west);
        \draw[->, thick] (cap7.east) -- (cap11.west);
        \draw[->, thick] (cap11.east) -- (cap8.west);
        \draw[->, thick] (cap8.east) -- (cap9.west);
        \draw[->, thick] (cap9.east) -- (cap12.west);
    \end{tikzpicture}
     \vspace{-1.0em}
    \caption{Execution of Cable Mounting Plan Generated by the LLM Planner.}
    \label{fig:exp_process}
    \vspace{-1.2em}
\end{figure*}

\textbf{Setup}
We use sigma.7 haptic devices~\cite{sigma7} to control two Franka Emika robots for collecting demonstrations.
The visual observation is captured by a third-person Intel RealSesne D435 camera.
As shown in Fig~\ref{fig:stretch_process} (a), tactile videos are collected by the GelSight ViTac sensor~\cite{yuan2017gelsight} on robot fingers.
F/T signals are collected by Bota Systems SensOne 6D force-torque sensor at the wrist, which also provides the human operator with haptic feedback during demonstration.
In the grounding stage, we use GPT-4 Omni as both analyzer and planner for its supreme compability of processing images and videos. 

\textbf{Evaluation Tasks}
We evaluate our framework on two sequential manipulation tasks, each presenting distinct challenges for task planning:
\begin{itemize}[wide]
    \item Cable mounting task, where a cable needs to be moved and inserted sequentially onto several clips.
    This task is a common process in industries like car manufacturing. 
    Inspired by how human handles this task, we use a bi-manual robotic setup. 
    One robot (called robot leader) holds the cable and moves it to each clip. 
    The second robot (called robot follower) joins the grasping when the cable reaches a clip, and both robots will perform the insertion together. 
    During demonstration, the operator teaches the robots to mount the cable onto two clips of different types (See Fig.~\ref{fig:stretch_process}(a) and task description in Fig.~\ref{fig:prompt_overview}(b)).
    For evaluation of planning compability, we randomize the number and position of clips as new task configurations.
    As for skills, we use implementations in our prior work~\cite{chen@contact}.
    \item Cap tightening task, where one robot should attach cap(s) onto bottles.
    During demonstration, the operator teaches the robot to pick a cap from the desk and tighten it to a target bottle.
    For evaluation of new task planning, the target bottle has an inner cap and an outer cap with randomized positions.
    The robot is supposed to tighten both caps to the target bottle.
\end{itemize}

\subsection{Evaluation on Demonstration Reasoning}\label{subsec: reason_eval}
Firstly, we evaluate the bootstrap reasoning pipeline in terms of its ability to extract correct skill sequences from demonstrations. 
We draw a comparison with control group A and B which rely solely on visual information to reason skill sequences.
The extracted skill sequence for cable mounting and cap tightening are presented in Fig.~\ref{fig: reason_comparison}.
While skills involving obvious movements (such as \smalltt{move\_object} and \smalltt{insert}) are successfully recognized by all groups, the LLM struggles to reliably identify skills involving intensive physical interactions (such as \smalltt{grasp} and \smalltt{stretch}) when only visual data is provided (Fig.~\ref{fig: reason_comparison} (a)).
Despite \smalltt{tighten} occurring multiple times in the cap tightening demonstration, the control groups manage to identify it only once (Fig.~\ref{fig: reason_comparison} (b)). Furthermore, \smalltt{tighten} is often misidentified as other skills, leading to an unreasonable sequence. 
This comparison highlights that incorporating object status data from tactile sensors in our pipeline significantly improves the accuracy and completeness of the skill sequences extraction.


Next, we evaluate the transition conditions reasoned by our pipeline using F/T signals.
Table~\ref{tab: skill_condition} shows \smalltt{stretched} and \smalltt{inserted} conditions for cable mounting as well as \smalltt{tightened} condition for cap tightening generated by the LLM, each containing force-relevant thresholds.
We observe that after F/T signals were introduced, the LLM retained its initial condition formulations but used the signals to update threshold estimations.
We then executed these skills on real-world robots to compare the success rate of each skill before (control group (c)) and after the update. 
Since directly assessing the success of the \smalltt{stretch} skill is challenging, we evaluated it in combination with the \smalltt{insert} process.
The \smalltt{insert} skill was tested on both clip types, with 20 trials conducted at random positions for each, while the \smalltt{tighten} skill was evaluated with the cap positioned at initial orientations ranging from 30 to 180 degrees. 
The success rates before and after the updates (shown in Table~\ref{tab: skill_condition}) indicate that the conditions for cable insertion into the U-type clip and cap tightening improved significantly with the integration of the demonstrated F/T signals. 
However, for the C-type clip which involves more dramatic and abrupt force changes, despite a rise in success rate after updates, the simple threshold reasoned by the LLM is still insufficient for robustly detecting success. 
More complex indicators, such as the change rate of resistance force, are likely required for better accuracy.

        


\subsection{Evaluation on Task Planning}
Finally, we evaluate the plans generated by the LLM for new task configurations. 
To compare the performance of our framework against all the control groups (a)-(d), we first assess the reasonableness of the generated skill sequences, assuming all skills in the plan will be executed successfully. 
Next, we test the plans on real-world robots, evaluating whether each skill is executed correctly without errors in the skill return (executability) and whether the entire task is completed successfully (success).
For instance, if the \smalltt{insert} success condition fails to detect when the cable has been inserted, the robots may continue pushing blindly, eventually throwing an error due to joint torque limits. 
In such cases, while the task itself is technically completed (the cable is inserted into the clip), the execution is considered a failure due to the error.

\begin{table}[t!]
\footnotesize
\caption{Evaluation on Skill Condition}
\centering
\resizebox{0.48\textwidth}{!}{
\begin{tabular}{p{0.95cm}|p{1.6cm}|p{0.8cm}|p{1.3cm}|p{0.8cm}}
 \toprule
 & \multicolumn{2}{|c|}{Before Update} & \multicolumn{2}{|c}{\textbf{After Update}} \\ 
 \hline
 & Condition & Success & Threshold & Success \\
 \hline
 Stretch & $f_r > 10\text{N}$ & \multirow{2}{*}{\centering 0.00} & $f_r > 9.5\text{N}$ & \multirow{2}{*}{\centering \textbf{0.90} } \\
 Insert U & $f_r < 5\text{N}$ &   & $f_r < 2.5\text{N}$ &  \\
 \hline
 Stretch & $f_r > 10\text{N}$ & \multirow{2}{*}{\centering 0.00} & $f_r > 9.5\text{N}$  & \multirow{2}{*}{\centering \textbf{0.50} } \\
 Insert C& $f_r < 10\text{N}$ &  & $f_r < 4.5\text{N}$ & \\
 \hline
 Tighten & $\tau_r>0.02\text{N}\cdot\text{m}$ & 0.00 & $\tau_r>2\text{N}\cdot\text{m}$  & \textbf{1.00} \\
 \bottomrule
\end{tabular}} 
\label{tab: skill_condition}
\end{table}

The overall performance is calculated as the average of the three criteria above. 
The evaluation results, presented in Table~\ref{tab: evaluation_task}, show that our framework outperforms all control groups, particularly in terms of task success rate.
One example of LLM-generated cable mounting plan as well as the execution process is demonstrated in Fig.~\ref{fig:exp_process}, which successfully accomplishes the assembly task.
For more detailed planning results, please refer to our accompanying video and project website.


\begin{table}[t]
\footnotesize
\caption{Evaluation on Task Planning}
\centering
\resizebox{0.48\textwidth}{!}{
\begin{tabular}{p{1.0cm}|c|c|c|c|c|c|c|c}
 \toprule
 & \multicolumn{2}{c|}{Reasonableness} & \multicolumn{2}{c|}{Executability} & \multicolumn{2}{c|}{Success} & \multicolumn{2}{c}{Overall} \\
 \hline
        & Cable & Cap & Cable & Cap & Cable & Cap & Cable & Cap \\
 \hline
\textbf{Ours}   & \textbf{1.00}   & \textbf{1.00}    & \textbf{0.91}  & \textbf{1.00}    & \textbf{0.91}  & \textbf{1.00}   & \textbf{0.94}  & \textbf{1.00} \\
 \hline
 Group A& 0.00   & 0.00 & 0.95  & 0.00 & 0.15 & 0.00 & 0.36 & 0.00 \\
 \hline
 Group B&  0.00   & 0.00 & 0.95  & 0.00 & 0.15 & 0.00 & 0.36 & 0.00 \\
 \hline
 Group C&  1.00 & 1.00 &1.00 & 1.00 &0.00  & 0.00 & 0.66 & 0.66 \\
 \hline
 Group D& 0.00 & 1.00 & 0.90 & 1.00 & 0.00 & 0.00 & 0.30 & 0.66 \\
 \bottomrule
\end{tabular}} 
\label{tab: evaluation_task}
\vspace{-2em}
\end{table}

\section{Conclusion}
We introduce an in-context learning framework that enables task planning for sequential, contact-rich manipulation tasks using multi-modal demonstration data. 
Our approach leverages tactile and force/torque information to segment and transform demonstrations into structured task plans, which then serve as reliable references for generating plans for new tasks.
While our method demonstrates effectiveness, its generalizability needs to be further validated across a broader range of manipulation tasks.
For future work, we plan to incorporate language instructions~\cite{zhou2023bridging} to further enhance LLMs' understanding of demonstrations. 
Additionally, fine-tuning a VLMto directly interpret tactile and force/torque data presents another promising approach to leverage multi-modal demonstrations.





\section*{ACKNOWLEDGMENT}
The authors acknowledge the financial support by the Bavarian State Ministry for Economic Affairs, Regional Development and Energy (StMWi) for the Lighthouse Initiative KI.FABRIK (Phase 1: Infrastructure as well as the research and development program under, grant no. DIK0249). 
The authors acknowledge the financial support by the Federal Ministry of Education and Research of Germany (BMBF) in the programme of "Souverän. Digital. Vernetzt." Joint project 6G-life, project identification number 16KISK002.

\bibliographystyle{IEEEtran}
\bibliography{references}

\balance

\end{document}